\documentclass{IOS-Book-Article}

\usepackage{mathptmx}
\usepackage{soul}\setuldepth{article}

\usepackage{amsmath,amssymb,amsfonts}
\usepackage{graphics}
\usepackage{graphicx}
\usepackage{cite}

\usepackage{algpseudocode}

\def\hb{\hbox to 11.5 cm{}}

\begin{document}

\pagestyle{headings}
\def\thepage{}
\begin{frontmatter} 

\title{
How the (Tensor-) Brain uses Embeddings and Embodiment to Encode Senses and Symbols
}

\markboth{}{September 2024\hb}

\author[A]{\fnms{Volker} \snm{Tresp}
\thanks{Corresponding Author: Volker Tresp, Ludwig-Maximilians-Universität München, 
 Oettingenstr. 67, D-80538 Muenchen, Volker.Tresp@lmu.de.}}
and
\author[A]{\fnms{Hang} \snm{Li}
}

\runningauthor{V. Tresp et al.}
\address[A]{LMU Munich}

\begin{abstract}













The Tensor Brain (TB) has been introduced as a computational model for perception and memory. This paper provides an overview of the TB model, incorporating recent developments and insights into its functionality. The TB is composed of two primary layers: the representation layer and the index layer. The representation layer serves as a model for the subsymbolic global workspace, a concept derived from consciousness research. Its state represents the cognitive brain state, capturing the dynamic interplay of sensory and cognitive processes. The index layer, in contrast, contains symbolic representations for concepts, time instances, and predicates. In a bottom-up operation, sensory input activates the representation layer, which then triggers associated symbolic labels in the index layer. Conversely, in a top-down operation, symbols in the index layer activate the representation layer, which in turn influences earlier processing layers through embodiment. This top-down mechanism underpins semantic memory, enabling the integration of abstract knowledge into perceptual and cognitive processes. A key feature of the TB is its use of concept embeddings, which function as connection weights linking the index layer to the representation layer. As a concept's ``DNA,'' these embeddings consolidate knowledge from diverse experiences, sensory modalities, and symbolic representations, providing a unified framework for learning and memory. Although the TB is primarily a computational model, it has been hypothesized to reflect certain aspects of actual brain function. Notably, the sequential generation of symbols in the TB may represent a precursor to the development of natural language. The model incorporates an attention mechanism and supports multitasking through multiplexing, simulating the brain's ability to rapidly switch between mental states. Additionally, the TB emphasizes multimodality, with the representation layer integrating inputs across multiple sensory and cognitive dimensions.

\end{abstract}

\end{frontmatter}
\markboth{September 2024\hb}{September 2024\hb}

\section{Introduction}

The debate between symbolic and subsymbolic processing in intelligent systems remains a central topic in cognitive science and artificial intelligence. The perspective presented here posits that symbols are the outcomes of measurement devices. These devices may be technical in nature or, as in this case, biological—representing the brain itself. Through the sensory system, the real world provides inputs to these measurement devices, with symbols serving as the resulting outputs. Importantly, these outputs can feed back into the processing pipeline, influencing subsequent operations. When the measurement device is an actor, such as a robot or a human, it can also actively modify the real world.

The Tensor Brain (TB), reviewed in this paper, builds upon these ideas. It suggests that while much of the brain's processing is subsymbolic, symbols play a crucial role at the highest levels of cognition.

A key component of the TB is the subsymbolic representation layer, which acts as a central communication hub akin to the ``global workspace'' in consciousness research. This layer functions as a “mental canvas” or “theater of the brain,” comparable to the blackboard concept in multi-agent systems. Brain modules can write to and read from this representation layer, making it a cognitive hub. Its activation state defines the cognitive brain state, encompassing the information that reaches conscious awareness and propagates throughout the system.

The second major component is the symbolic index layer, where each symbol or index has a localized representation. For example, when the brain perceives a scene and identifies an entity as a dog, it acts as a bottom-up measurement device, producing the symbol \textit{Dog} as the output. Symbols and subsymbolic representations interact, creating complementary descriptions of the world. Every symbol activation is associated with subsymbolic activity, while subsymbolic representations are continuously annotated with symbolic labels.

The connection weights between the symbolic index and the representation layer form the embedding vector of the symbol. This embedding, in a way a  symbol's ``DNA,'' integrates perceptual inputs, experiences, and embodied knowledge. When a symbol like  \textit{Dog} is activated—whether due to sensory input or deliberate focus—it triggers activity in the representation layer via its embedding, propagating this information to other brain regions through grounding and embodiment. Symbols also interact with other symbols, forming networks of relationships. The embedding vectors are optimized to fulfill their roles in this network.

\begin{figure}[t]
\begin{center}
 \includegraphics[width=\linewidth]{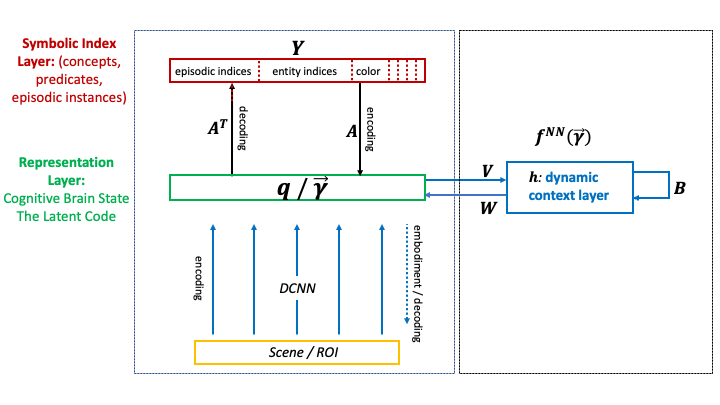}
\end{center}
\caption{The Tensor Brain (TB) Architecture: The architecture processes scene input (illustrated at the bottom) through the layers of a deep convolutional network, mapping it to the representation layer, which serves as a mathematical model of the global workspace.
On the right, the evolution neural network, featuring one hidden layer, provides recurrence. Within the TB framework, this recurrent component is referred to as the dynamic context layer.
The representation layer connects to the index layer, which in turn feeds back into the representation layer. The embedding vectors are represented by the columns of matrix 
$\mathbf{A}$.
Through bottom-up and top-down processing, the system can generate multiple labels for a given scene or region of interest (ROI).
}
\label{fig:TB-arch}
\end{figure}

Figure~\ref{fig:TB-arch} illustrates the TB architecture, showing how the representation layer integrates sensory inputs and interacts with symbols through the index layer. The activation state of the representation layer serves as a distributed code, providing a latent representation for both sensory input and symbolic indices.

\begin{figure}[t]
\begin{center}
 \includegraphics[width=\linewidth]{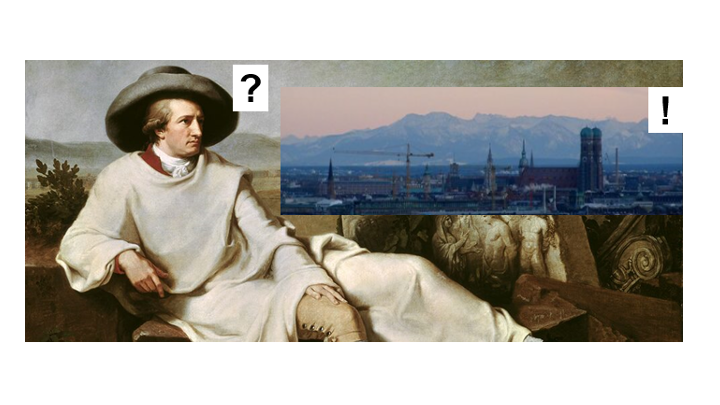}
\end{center}
\caption{
Johann Wolfgang von Goethe gazes upon a picturesque landscape, with mountains forming the distant backdrop (perception). As his eyes sweep across the scene, he identifies the Frauenkirche (symbolic labeling) and deduces that the city before him must be Munich (semantic memory). Observing the mountains, he further concludes they must be the Alps (semantic memory).
This view triggers memories of his last visit to Munich (episodic memory), particularly a delightful dinner at the Flaucher. Reflecting on the weather forecast, which promises a pleasant evening, Goethe contemplates whether to revisit the Flaucher for another memorable meal (decision support through future episodic memory and imagination).
}
\label{fig:TB-Goethe}
\end{figure}

In the TB framework, symbols represent concepts such as entities, classes, attributes, predicates, and episodic instances. This enables the formation of episodic memories, where symbolic similarity contributes to memory recall. The similarity between two scenes, for instance, depends not only on sensory input but also on their symbolic representations. This dynamic interplay between perception and memory reflects Goethe’s adage: ``You only see what you know!'' Figure~\ref{fig:TB-Goethe} demonstrates how episodic and semantic memory support perception.

Symbolic decoding—essential for perception and memory—is conceptualized as a serial process, akin to an ``inner thought language.'' Humans excel at articulating perceptions and memories with minimal effort, suggesting that spoken language evolved from an earlier, internal symbolic language. While much reasoning and communication rely on symbols, the cognitive brain state encompasses more than just symbolic activity. It is reflected in embodiment and grounding, manifested through intonation, gestures, body language, and facial expressions.

The TB framework aligns with the notion of the brain as a prediction machine, a view supported by substantial evidence at the level of implicit memories, such as perceptual and motor skills. However, to make predictions over longer timescales—such as planning an evening—the mind must relate the present to both past experiences and future possibilities. This requires explicit understanding, explicit memory, and imagination.

The remainder of the paper is organized as follows.
Section 2 summarizes the literature on the Tensor Brain.
Section 3 introduces the representation layer as a model for the global workspace and defines the cognitive brain state, illustrating its progression over time as a recurrent neural network.
Section 4 provides a probabilistic interpretation of the cognitive brain state.
Section 5 discusses symbolic decoding as bottom-up inference, covering symbolic indices for concepts, predicates, and episodic time instances.
Section 6 explains symbolic encoding as top-down inference, introducing an attention mechanism and discussing embodiment as an encoder in an autoencoder framework.
Section 7 describes the Tensor Brain's operational modes, including perception, episodic memory, and semantic memory. It elaborates on the generation of triple statements and language understanding.
Section 8 details how self-supervised learning adapts embedding vectors.
Section 9 examines multitasking via multiplexing and cognitive control.
Section 10 explores multimodality and various forms of reasoning, including embedded reasoning, symbolic reasoning, and embedded symbolic reasoning.
Section 11 concludes with a summary of key findings and implications.

 \begin{figure}[t]
\begin{center}
 \includegraphics[width=\linewidth]{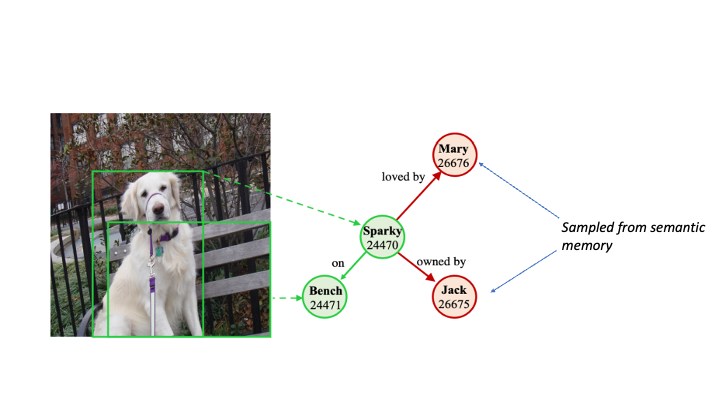}
\end{center}
 \caption{
Experimental results from the Tensor Brain (TB) for perception, enhanced by semantic memory, are shown in \cite{tresp2023tensor}. In the first region of interest (ROI), the left bounding box is identified as \textit{Sparky}. The top-ranked labels for this ROI include: \textit{Dog}, \textit{Mammal}, \textit{LivingBeing}, \textit{Young}, \textit{White}, and \textit{OtherActivity}. For the second ROI, the top-ranked labels are: \textit{Bench}, \textit{Furniture}, \textit{NonLivingBeing}, \textit{Old}, \textit{OtherColor}, and \textit{OtherActivity}.
The sampled binary statements are: \textit{(Dog, sit on, Bench)}, \textit{(LivingBeing, on, Furniture)}, \textit{(Mammal, sit on, Old)}, \textit{(White, sit on, Bench)}.
Additionally, semantic memory enhances perception by providing binary statements related to entities \underline{not present in the scene}, such as: \textit{(Sparky, ownedBy, Jack)} and \textit{(Sparky, lovedBy, Mary)}, where \textit{Jack} and \textit{Mary} are part of the agent's semantic memory, but not visible in the scene.
 }
\label{fig:TB-dogench}
\end{figure}

\mbox{}\\
\noindent \textit{Glossary and notation:}
\begin{itemize}
\item 
The \textbf{agent} represents an \textbf{individual}  and serves as the actor, with its \textbf{mind} executing cognitive functions through its neurobiological foundation—the \textbf{brain}.

\item The \textbf{representation layer} is a model for the brain's \textbf{global workspace}. Its activation is the \textbf{cognitive brain state} (CBS). The probabilistic interpretation of the CBS is the \textbf{probabilistic cognitive brain state} (pCBS).

\item The \textbf{index layer} encodes indices which can be concepts, predicates, and episodic indices. Each index is realized by an 
\textit{ensemble of neurons}.
 An ensemble is referred to as an \textbf{index}, a \textbf{symbol}, or a \textbf{label}.

\item 
The synaptic weights linking an index with the representation layer form the \textbf{embedding} vector of that index. 
An embedding vector is the signature or ``DNA'' of the associated index.

\item 
$\vec \gamma$ is the vector of post-activations of the ensembles in the representation layer, 
and $\mathbf{q}$ is the vector of pre-activations. $\mathbf{a}_k$ is the embedding vector of index $k$. Components are 
$\gamma_i$, 
${q}_i$, and ${a}_{i, k}$, with $ i = 1, \ldots n$.

\end{itemize}

\section{The Development of the Tensor Brain}
\label{sec:evolTB}


This section provides an overview of the key milestones in the development of the Tensor Brain (TB) model, tracing its evolution and contributions to the fields of perception, memory, and cognitive modeling.

The first foundational paper on the TB model was published in 2015 \cite{tresp2015learning}. It explored the interaction between symbolic representations, subsymbolic representations, and concept embeddings in brain function. The paper introduced the Tucker tensor decomposition—a mathematical approach for reducing the complexity of multi-dimensional data—as a generative model for perception and memory processes. Central components of the TB model, including the representation layer and the index layer, were introduced.

The work connected semantic memory to knowledge graph factorization and episodic memory to the factorization of temporal knowledge graphs, pioneering the field of temporal knowledge graph embedding models. Additionally, it proposed the analysis of visual scenes through factorized graph models, which later evolved into the scene graph embedding models widely used today. The paper also renewed interest in index-based approaches to memory, echoing the work of Teyler and colleagues \cite{teyler1986hippocampal}.

The vision component of the TB model was extended in \cite{baier2017improving} by incorporating bounding boxes as regions of interest (ROIs) for identifying entities within a scene. This extension included experimental validation using annotated image data from \cite{johnson2015image}. Concurrently, the concept of temporal knowledge graphs received further exploration in works such as \cite{tresp2017embedding} and \cite{ma2018embedding}, solidifying its role in the model’s development.

In 2020, \cite{tresp2020tensor} introduced a probabilistic generative model for the TB, enriching its theoretical underpinnings. Meanwhile, \cite{sharifzadeh2021classification} focused on integrating an attention mechanism and emphasized the role of prior knowledge and grounding in the model.

The TB model underwent further refinement in \cite{tresp2023tensor}, which integrated extensive experimental evidence and provided a robust mathematical framework. This work established critical connections between the TB approach and cognitive neuroscience, enhancing its relevance for both theoretical and applied research.
The most recent development, \cite{tresp2024QTB}, derived a probabilistic Tensor Brain model inspired by the Heisenberg measurement process, drawing parallels to concepts in decoherent quantum theory.

\section{The {Subsymbolic} Representation Layer}
\label{sec:rl}

\subsection{The Representation Layer and the Cognitive Brain State (CBS)}

 The brain is composed of thousands, potentially millions, of interconnected modules that must coordinate and exchange information effectively. To support higher-order processing, including conscious awareness, an integrative mechanism is essential for organizing and relaying information across these modules.

In the tensor brain (TB) model, the representation layer serves as an integrative, high-dimensional workspace where modules can exchange information by writing to and reading from shared representations. This layer facilitates the coordination of neural activities and synthesizes the cognitive brain state (CBS).

The CBS is defined by the activation pattern of all ensembles in the representation layer. Specific ensembles may represent concrete features, such as the color 'red,' while others encode abstract latent factors. Experimental findings suggest that the CBS exhibits a well-organized clustering structure, reflecting meaningful groupings of cognitive and perceptual representations \cite{tresp2023tensor}.

Let $n$ represent the total number of neuron ensembles in the representation layer. The activation $\gamma_i$ of each ensemble $i$ is calculated as a function of its pre-activation input $q_i$ as
\begin{equation} \label{eq:preactact}
\gamma_i \leftarrow \mathrm{sig} (q_i ).
\end{equation}
Here, $\gamma_i \in [0, 1]$ denotes the post-activation value, where $\gamma_i = 1$ indicates maximal firing intensity, and $\gamma_i = 0$ corresponds to basal or no firing. The function $\mathrm{sig} (q_i) = (1 + \exp(-q_i))^{-1}$ is the standard logistic activation function.

In this framework, $q_i$ represents the cumulative input signals received by ensemble $i$, while $\gamma_i$ represents the corresponding output activation. This dynamic ensures that the representation layer accurately reflects ongoing cognitive states. The CBS, represented by the activation vector $\vec{\gamma}$ of the representation layer, encapsulates the brain's distributed coding of sensory inputs and higher-order representations. This framework enables the TB model to simulate cognitive processes such as perception, memory integration, and decision-making.

\subsection{The Representation Layer, the Global Workspace Theory, and the 
Blackboard}
 
 
Baars and his colleagues introduced the concept of a global workspace\cite{baars1997theater}, which serves as the foundation for several theories of consciousness\cite{baars1997theater, dehaene2014consciousness}. The global workspace functions as a central hub for broadcasting and integrating information, enabling its dissemination across various brain modules. Often referred to as the ``theater of the brain'' or the ``mental canvas,'' the global workspace has been likened to a blackboard in a multi-agent system, where computational modules collaborate by sharing information~\cite{baars1997theater}.

The global workspace facilitates communication and interaction, particularly among modules that handle higher levels of abstraction, where direct, module-to-module interactions may be insufficient or impractical.


Consider the example of vision. Visual inputs are processed through visual pathways, contributing to the pre-activations of neural ensembles within the global workspace. These inputs are ``written'' into the representation layer, forming part of the cognitive brain state (CBS). In turn, the CBS can feed back into the visual pathways. This feedback, often associated with embodiment, acts as an input to the early visual processing layers, modulating their activity. Thus, the global workspace enables a bidirectional exchange of information: visual input informs the global workspace, and the global workspace influences early visual processing.

Beyond vision, the global workspace integrates signals from a wide array of brain modules, each corresponding to different sensory modalities, such as vision, hearing, taste, smell, touch, proprioception, and pain. Crucially, the global workspace is not merely a passive integrator. Brain modules can also be affected by the state of the global workspace, creating a dynamic feedback loop.

This process extends to states of consciousness. It has been proposed that all brain states capable of reaching consciousness—such as inner feelings, emotional states, and even pain after injury—contribute to the global workspace. In turn, these states can be influenced by the global workspace's current configuration. However, not all brain modules have a direct impact on the global workspace. Consequently, some processes or information may never reach conscious awareness.

While sensory input plays a dominant role in shaping the global workspace, other factors—such as emotions tied to memories or physical pain—also contribute significantly. For example, the discomfort from an injury or the emotional resonance of a recalled memory can profoundly affect the global workspace's state.

%
%
 
 In the tensor brain (TB) model, the global workspace is hypothesized to be implemented by the representation layer. This representation layer is conceptually related to the hidden layer of a recurrent neural network (RNN), but with critical modifications to accommodate the unique requirements of the TB architecture. These modifications are explored in greater detail in the following discussion.

\subsection{Evolution Neural Network and Recurrency}

%
%
 The representation layer in the tensor brain (TB) model incorporates feedback connections, creating dependencies between the activations of neural ensembles and an internal memory. This feedback mechanism is modeled as follows,
 for $i=1, \ldots, n$
 \begin{equation} \label{eq:tiev}
q_i^{(\tau)} \leftarrow 
q_i^{(\tau-1)} + 
g_i (\mathbf{v}^{(\tau)} ) + 
 f_i^{\textit{NN}}
 (\vec \gamma^{(\tau-1)} ) .
 \end{equation}
 Here, 
$\tau$ denotes a discrete-time step, representing sequential operations of the brain. All computations performed during a specific 
$\tau$  are linked to the same concept or episodic index.

 THen, $f_i^{\textit{NN}}
 (\vec \gamma^{(\tau-1)} )$ is the $i$-th output of a neural network with linear output units 
 This component, known as the evolution neural network, models interactions among ensembles and their evolution over time. 
 $\mathbf{v}^{(\tau)} $ 
 represents an external input (e.g., from a visual scene). This input is processed by a deep convolutional neural network, 
 $\mathbf{g} (\mathbf{v}^{(\tau)} )$, which maps sensory input to the representation layer using linear output units.
%
%
 Recall that 
 $ \gamma^{(\tau)}_i \leftarrow 
 \mathrm{sig} (q^{(\tau)}_i )$
 (Equation\ref{eq:preactact}). 
 
 In this framework, a specific 
$q_i^{(\tau)} $ 
  can depend on only a subset of ensembles, promoting modularity in the system.
The term 
$q_i^{(\tau-1)} $, representing the state of the ensemble at the previous time step, introduces a skip connection (a common feature in deep learning models like ResNets~\cite{he2016deep} and LSTMs~\cite{hochreiter1997long}).\footnote{The skip connection was not used in \cite{tresp2023tensor}} This dependency implements self-memory, allowing the system to maintain continuity across time steps.

The evolution neural network enables the prediction of future cognitive brain states (CBS), a critical capability for tasks involving sequential reasoning or anticipating future events. By modeling relationships in both semantic memory (knowledge about the world) and episodic memory (personal experiences), the network supports the integration of knowledge and experience into the CBS. This dual functionality is central to the TB model’s ability to handle complex cognitive tasks.

 \section{The Probabilistic Cognitive Brain State (pCBS)}
 \label{sec:pcbs}
 
 The post-activation $\gamma_i$ can be interpreted as the parameter of a Bernoulli distribution, where
 \begin{equation}
 P(X_i = 1) = \gamma_i .
 \label{eq:noise}
 \end{equation}
 Here, $X_i = 1$ represents the ensemble $i$ being “on,” while 
 $X_i = 0$ represents it being “off.”
Assuming mutual independence among ensembles, the probabilistic cognitive brain state (pCBS) can be defined as a joint probability distribution
 \[
P(X_1 = i_1, ..., X_n = i_n) = 
\prod_{j=1}^n (\gamma_j)^{i_j} (1-\gamma_j)^{1-i_j}
 \]
 where $i_j \in \{0, 1\}$. 
This probabilistic formulation of the CBS introduces a new layer of interpretability and flexibility, as extensively discussed in  \cite{tresp2024QTB}. 
 It supports top-down inference and extends the model to account for uncertainty and variability in neural activation patterns.\footnote{A distribution with maximum entropy corresponds to $\gamma_j = 1/2, \forall j$, which implies that, in the absence of input, all neurons fire with half intensity. While this may not be physiologically plausible for real neurons (e.g., neurons with ReLU activation functions remain inactive without input), it reflects the statistical nature of the pCBS.
  }

 Equation~\ref{eq:tiev} now has two interpretations. First, one can assume that the dynamics of the neural activations is the basic deterministic dynamic equation, and Equation~\ref{eq:noise} describes a noisy measurement. 
 Alternatively, one can consider the states of the random variables $X_i$ to be the fundamental quantity and we need to consider the evolution of a multivariate probability distribution. 
In \cite{tresp2024QTB}, it is demonstrated that the evolution neural network can be derived from the latter probabilistic perspective, assuming certain approximations and independence constraints. Specifically, the pCBS is approximated at each instance by 
$n$ independent Bernoulli distributions.
While the brain’s physical reality is represented by the firing rates 
$\gamma_i$, the random variables 
$X_i$
  serve as mathematical constructs for modeling purposes and are not directly observed.
  

\section{The Symbolic Index Layer}
\label{sec:symbenc}

\subsection{The Index Layer}

The symbolic index layer is a cognitively significant module that interacts dynamically with the representation layer. It not only receives input from the representation layer but also provides feedback, establishing a bidirectional connection.

In the brain, an index (or pointer) may be represented by an ensemble of neurons, similar to other neural ensembles. However, a distinguishing feature of the index layer is its ability to enforce a sample-take-all mechanism—a variation of the well-known winner-take-all concept.

Each index corresponds to a distinct symbol, and we categorize these symbols into three types: concept indices, predicate indices, and episodic indices. Importantly, we propose that an episodic index can also be treated as a symbol. In perceptual tasks, the term label is  used interchangeably with  symbol.

%
%
%
%

 \subsection{Concept Indices}
 \label{sec:conc}


Indices associated with concepts are inherently symbolic. These concepts can represent a wide range of entities, such as objects, classes, attributes, locations, inner emotional states, actions, or decisions. Concept indices enable the recognition and encoding of stable patterns within the Cognitive Brain State (CBS). For example, a cluster analysis of the CBS could allow the brain to detect recurring patterns corresponding to concepts like \textit{Dog}, a specific dog named \textit{Sparky}, or attributes such as \textit{Black} and \textit{Happy}.

In perception, concept indices may be used to label scenes or define regions of interest (ROIs) within a visual field. This labeling mechanism is essential for identifying areas that are contextually or semantically significant.

Assuming the CBS is represented by the activation vector $\vec \gamma$, 
 an index $k$ in the index layer receives input from the representation layer, with its activation governed by
 \begin{equation} \label{eq:conlabel}
P(Y=k | \vec \gamma) 
=
 \mathrm{softmax}_{\textit{dom}} (a_{0, k} + 
\sum_{i=1}^n
 a_{i, k} \gamma_i ) .
\end{equation} 
Here,  $Y =k $ indicates that the index layer is in state $k$, i.e., 
only index 
$k$ is active, while all other indices remain inactive. 
The vector $\mathbf{a}_k = (a_{1, k}, \ldots, a_{n, k})^\top$ is the typically sparse embedding vector of concept $k$, representing its connection weights to the representation layer. 
$a_{0, k} $ is a bias term to concept $k$. 
In the architecture, $a_{i, k}$ is the weight on the link from ensemble $i$ in the representation layer to ensemble $k$ in the index layer.
The softmax function is defined as 
\[
 \mathrm{softmax}_{\textit{dom}} (a_{0, k} + 
\sum_{i=1}^n
 a_{i, k} \gamma_i )
 = 
 \frac{
 \exp (a_{ 0, k} + 
\sum_{i=1}^n
 a_{i, k} \gamma_i )
 }
 {
\sum_{k' \in \textit{dom} } \exp (a_{ 0, k'} + 
\sum_{l=1}^n
 a_{l, k'} \gamma_l ) 
 }.
\]
It normalizes over a domain (\textit{dom}) such as the set of all entities, predicates, classes, or colors.
Each domain defines a mutually exclusive and collectively exhaustive set of labels, although in practice, violations such as assigning multiple labels (e.g., two colors) to the same entity might occur. See Section~\ref{sec:fcp} for further discussion.\footnote{For the pCBS, Equation~\ref{eq:conlabel}
approximates
 $P(Y=k | \textit{context}) = \mathbb{E}_{P(X_1, ..., X_n | \textit{context})} P(Y =k | X_1, ..., X_n) $~\cite{tresp2024QTB}. }

 
 
 Concept indices capture repeated patterns within the CBS and are identified through their embeddings or potentially their location within a topographic map. While these indices in animals do not correspond to explicit names like
 \textit{Dog}, \textit{Sparky}, \textit{Black} and \textit{Happy}, in humans, they directly reflect concepts encoded in natural language.

 \subsection{Predicate Indices}

 
 The second type of indices refers to predicates. Imagine an agent analyzing a scene. The CBS first identifies a region of interest (ROI) that, for instance, contains 
  \textit{Sparky}, a \textit{Dog} in the scene.
   Next, it identifies a second ROI, which, for example, contains \textit{Jack}, a \textit{Person}, in the scene. A third ROI is then formed that encompasses both of the previous ROIs. Within the context of these two ROIs, the third one may be labeled according to their relationship, such as \textit{looksAt}. Therefore, the brain not only considers the labels of individual ROIs, but also the labels of the relationships between them. In this context, $\mathbf{a}_k$ represents the embedding vector of predicate $k$.
  
%

\subsection{Episodic Indices Refer to Time Instances}
\label{sec:timein}

 The third type of indices pertains to episodic indices and relates to time instances. In the TB, an episodic index is introduced for each relevant time instance. In its simplest form, the embedding vector of index $t$ corresponds to the vector of pre-activations from the representation layer at that specific instance, i.e., $\mathbf{a}_t \leftarrow \mathbf{q}^{(t)}$. If the representation layer is primarily influenced by visual input from a scene, it will store the resulting pre-activations of the CBS, which may include information such as the emotional state at that time. Later, we will explore how $\mathbf{a}_t$ is optimized.


 \section{Symbolic Encoding}
 \label{sec:symbdec}
 
 

 The indices or symbols in the index layer can be encoded into subsymbolic representations, a process that embodies top-down inference. The primary purpose of top-down inference is to inform the representation layer—and, by extension, the entire brain—about which concept has been detected in vision, or more generally, which concept the brain is currently focusing on. For instance, if the concept of \textit{Sparky} is detected, this information should be communicated to the representation layer and, ultimately, to the brain. This is particularly important for integrating memory with perception. With the current hypothesis that \textit{Sparky} is present in the image, the brain can augment this perception with background knowledge, such as that \textit{Sparky} is \textit{Friendly}, \textit{ownedBy} \textit{Jack}, and \textit{lovedBy} \textit{Mary} (see Figure~\ref{fig:TB-dogench}). We will discuss four variants of this process.

 First, we explore the deterministic model without top-down inference. Second, we examine how a sampled concept label feeds back into the representation layer. Third, we address the generation of multiple sampled concepts. Finally, we discuss the attention approximation.

 \subsection{A Deterministic Model without Top-down Inference}
 

In a deterministic CBS interpretation, Equation~\ref{eq:conlabel} describes the system's output. Suppose the labels for \textit{Sparky}, \textit{Dog}, \textit{Black}, and \textit{Happy} show substantial activation. The agent might select the label with the highest probability, such as \textit{Sparky}. Alternatively, the agent could consider other likely labels, such as \textit{Dog}, \textit{Black}, and \textit{Happy}. In either case, there is no feedback to the representation layer, and the rest of the brain remains uninformed about the likely labels. This setup reflects the standard RNN model, where all labels are treated as independent. However, this is not the approach used in the TB.
 
 \subsection{A Single Sampled Label}

Assume that the labels for \textit{Sparky}, \textit{Dog}, and \textit{Black} exhibit substantial activation. In sampling mode, the TB samples from the distribution in Equation~\ref{eq:conlabel} and may generate the label \textit{Sparky}. Generally, when index $k$ is sampled, it is assumed that concept $k$ is represented in the representation layer.

The index layer operates in a ``sample-take-all'' mode, meaning only one index fires at a time, suppressing the firing of all other indices. In the example, the index \textit{Sparky} would be the active one. Technically, after sampling, the state of the index layer is represented as a one-hot vector.

When index $Y = k$ fires, it activates the global workspace in a top-down manner according to the following update rule
 \begin{equation} \label{sec:stateup}
\mathbf{q} \leftarrow \alpha \mathbf{q} + \beta \mathbf{a}_{k} .
\end{equation}
Here, $\mathbf{a}_k$ represents the embedding vector for concept $k$. The term $a_{i,k}$ can be interpreted as the synaptic weight connecting ensemble $k$ in the index layer to ensemble $i$ in the representation layer. In the pCBS interpretation, we obtain independent Bernoulli distributions. This update equation was derived in \cite{tresp2024QTB}.

The factor $\alpha$, where $0 \le \alpha \le 1$, controls how much of $\mathbf{q}$ is preserved during the update. With $\beta = 0$ and $\alpha = 1$, we recover an RNN update, where the CBS is unaffected by the sample. With $\beta = 1$ and $\alpha = 0$, we obtain the Heisenberg approximation, where the CBS is entirely overwritten by the sample's embedding vector. Finally, with $\beta = 1$ and $\alpha = 1$, we have the Heisenberg approximation with an ``added prior.'' This latter version is used in the TB.

As reflected in the last equation and in Equation~\ref{eq:conlabel}, the connections are bipartite and symmetrical. Strict symmetry is implied by some theoretical considerations~\cite{tresp2024QTB}, though it is not a strict requirement and is likely to be violated biologically. Note that when $\alpha = 1$, the previous pre-activation is not eliminated but rather added to the embedding vector, acting as a skip connection.

\subsection{Multiple Sampled Labels}
 
 The process can be repeated multiple times by continuously sampling from Equation~\ref{eq:conlabel} and updating Equation~\ref{sec:stateup}. With $\alpha > 0$, each sample is always conditioned on the previous ones. Each sample represents a specific interpretation of the CBS, framed within the context of prior samples. For instance, the first sample might correspond to the label \textit{Sparky}, followed by \textit{Dog}, then \textit{Black}, \textit{Happy}, and so on.

Finally (with $\beta = 1$ and $\alpha = 1$, as in the TB), we have the following update rule
 \begin{equation} \label{sec:stateup22}
\mathbf{q} \leftarrow \mathbf{q}+ \sum_{k \in \textit{Sample}} \mathbf{a}_{k}.
\end{equation}
Thus, the index layer functions as part of a labeling engine. This clearly contrasts with a standard RNN, where the last equation would not include the summation term—meaning there is no feedback from the samples to the state. In the TB, the representation layer is explicitly informed about all generated samples, and subsequent samples are generated with consideration of all previously generated labels.

The activation of an index is an internal brain measurement that feeds back into the system, allowing all modules in the brain to learn about what has been detected. Equation~\ref{sec:stateup2} illustrates the additive combination of embedding vectors, a concept found in models like word2vec \cite{mikolov2013distributed} and TransE \cite{bordestranslating2013}.
 
%

 \subsection{Probabilistic Averaging: Attention}


Consider the case where there is no sampling, and the CBS is updated by taking into account all possible labels, weighted by their probabilities. In \cite{tresp2023tensor,tresp2024QTB}, the following equation is introduced
 \begin{equation} \label{sec:stateup2}
\mathbf{q} =
\mathbf{q} + 
\sum_{k} \mathbf{a}_{k} \mathrm{softmax}_{\textit{dom}} (a_{0, k} + 
\sum_{l=1}^n a_{l, k} \gamma_l ) .
\end{equation} 
This equation provides an update to the CBS, not through sampling, but by considering the probabilities of the labels. It can be interpreted as a form of attention mechanism with a skip connection, commonly found in generative AI models. In this context, the CBS vector $\vec{\gamma}$ acts as the query vector, while the embedding vectors $\mathbf{a}_k$ (for all $k$) serve as both the key and value vectors.

 \subsection{Attention versus Sampling}

Attention does not yield a symbolic interpretation because no single index is active in isolation. However, a key advantage of the attention mechanism is its speed, parallelism, and holistic consideration of all possible interpretations. The attention equation links indices to related indices, facilitating the sharing of statistical strength across them.

In \cite{tresp2024QTB}, it is shown that the attention mechanism can be derived from a Heisenberg framework rather than a Bayesian one. In a Bayesian scenario, only the indices that are actually sampled would influence the CBS, while in the Heisenberg scenario, all indices are considered together.

Sampling commits the brain to a specific interpretation. For instance, if the black object in a scene is sampled as \textit{Sparky}, the brain can incorporate a wealth of background information about \textit{Sparky}. However, if the object is sampled as a \textit{Puma}, the brain adds different background knowledge, and the agent’s response may differ significantly. Sampling allows both possibilities to be explored in separate sampling rounds, enabling the exploration of joint dependencies between labels. This process of sampling and top-down inference forms the foundation of semantic memory, as illustrated in Figure~\ref{fig:TB-segm}.

The sampling of indices is a fundamental feature of the TB. But what triggers the sampling? Sampling appears to share many similarities with the collapse of the wave function in quantum mechanics. Does this suggest that the brain functions as a decoherent quantum computer? These questions are explored in \cite{tresp2024QTB}. Additionally, \cite{wong2023word} discusses sequential Monte Carlo (SMC) as a sampling method. In contrast to SMC, the TB does not require the filtering step because the labels are self-generated and do not depend on external measurements. 
Sensory data provides deterministic inputs and are ``given.''

 \subsection{Attention, Autoencoding and Embodiment}
 \label{sec:auto}

 We examine two types of autoencoders where the representation layer functions as the latent code or bottleneck layer.

In the first autoencoder, indices represent the visible data points, while the cognitive brain state (CBS) serves as the latent code. The decoder maps the CBS to the index layer activation (Equation \ref{eq:conlabel}), where an index is sampled. The  sample forms the data point. The encoder operates in the reverse direction, mapping an index to the representation layer via Equation \ref{sec:stateup}, with parameters $\beta = 1$ and $\alpha = 0$.

In the second autoencoder, sensory inputs constitute the visible representation. The encoder maps visual input $\mathbf{v}$ to the index layer using a visual processing function $\mathbf{g}(\mathbf{v})$. The decoder reconstructs the visual input from the CBS by approximating the inverse of the visual processing function, i.e., $\mathbf{\hat v} = \mathbf{g}^{\text{inv}}(\vec{\gamma})$. This top-down processing reflects a form of grounding or embodiment: it not only updates the representation layer but also propagates changes through all visual processing layers to reflect the consideration of the concept \textit{Sparky} in the index layer.

In brain circuits, connections are typically bi-directional—if one module connects to another, there is usually a reciprocal connection. Consistently, the TB model assumes bi-directional connections between the representation layer and the index layer, as well as within the perceptual processing pipeline. However, the inverse perceptual pipeline often lacks precision. For example, while observing \textit{Sparky} in a scene provides a vivid and detailed image, a mental reconstruction of \textit{Sparky} is generally less clear and detailed.

\section{Operational Modes}
\label{sec:opmodes}


One of the strengths of the TB approach is its ability to model different operational modes of the brain within a single architecture. Specifically, we explore how the TB can represent perception, episodic memory, and semantic memory.

\subsection{Pseudocode}

The pseudocode is described in Algorithm~\ref{alg:cap-1}, Algorithm~\ref {alg:cap-22}, 
and Algorithm~\ref {alg:cap-33}.

\begin{algorithm} [h]
\caption{Evolution Neural Network 
\\ \mbox{} }
\label{alg:cap-1}
\begin{algorithmic}[1]
\State Input: $\mathbf{q}^{\textit{in}}$ \Comment{The CBS is the input} \label{algo: 1}
\State $q_i \leftarrow 
q_i^{\textit{in}} + 
 f_i^{\textit{NN}}
 (\mathrm{sig} (\mathbf{q}^{\textit{in}})), \forall i$ \Comment{Evolution NN with skip connection} \label{algo: 2}
 \State Output: $\mathbf{q}$
\end{algorithmic}
\end{algorithm}

\begin{algorithm} [h]
\caption{Input and Attention 
\\ \mbox{} }
\label {alg:cap-22}
\begin{algorithmic}[1]
\State Input: $\mathbf{q}, \mathbf{v}$ \Comment{The CBS is the input} \label{algo: 11}
 \State $\mathbf{q}\leftarrow \mathbf{q}+ \mathbf{g}(\mathbf{v} )$ \Comment{
 $\mathbf{v}$ from scene or bounding box} \label{algo: input}
 \State $\mathbf{q}
\leftarrow \mathbf{q}+ 
\sum_{k} \mathbf{a}_{k} \mathrm{softmax}_{\textit{dom}} (a_{0, k} + 
\sum_{l=1}^n a_{l, k} 
\mathrm{sig} (q_l) ) $ \Comment{Attention} \label{algo: att}
 \State Output: $\mathbf{q}$
\end{algorithmic}
\end{algorithm}

\begin{algorithm} [h]
\caption{Decoding (bottom-up) and encoding (top-down); the Heisenberg approximation with an added prior (used in the TB) uses $\alpha= 1$ and $\beta= 1$
\\ \mbox{} }
\label {alg:cap-33}
\begin{algorithmic}[1]
\State Input: $\mathbf{q}$, $\alpha$, $\beta$ \Comment{The CBS is the input} \label{algo: 12}
 \State Sample $k \sim 
\mathrm{softmax}_{dom} (a_{0, k} + 
\sum_{l=1}^n a_{l, k} 
\mathrm{sig} (q_l) )$ \Comment{Bottom-up: $k$ is sampled} \label{algo: 4}
 \State$\mathbf{q} \leftarrow \alpha \mathbf{q} + \beta \mathbf{a}_k $ \Comment{Top-down inference} \label{algo: tdstep}
 \State Output: $\mathbf{q}, k$
 \end{algorithmic}
\end{algorithm}
%


\subsection{Visual Perception}

\subsubsection{Visual Scene}


Assume the agent is analyzing a visual scene. Algorithm~\ref{alg:cap-22} begins with an initial neutral input, such as $\mathbf{q} = 0$. The visual input, $\mathbf{v} \leftarrow \mathbf{v}^{\textit{scene}}$, is then processed (Line~\ref{algo: input}). Next, the attention mechanism is applied to extract relevant information from past episodic memories (Line~\ref{algo: att}).

Algorithm~\ref{alg:cap-33} describes the decoding and encoding process. A label is sampled from the scene (bottom-up), and the sampled label is fed back to the representation layer (top-down). Algorithm~\ref{alg:cap-33} may be called multiple times, with each iteration generating a label $k$. These labels might, for instance, indicate the agent's location (e.g., \textit{EnglishGarden}) or the weather condition (e.g., \textit{Sunny}). Additionally, a generated label could represent a past time-index, suggesting a similarity to a previous scene.

\subsubsection{A First ROI in a Visual Scene}

%
 The agent may then focus on the properties of an entity, which are typically defined by a specific region of interest (ROI) in the scene. The TB applies the evolution neural network from Algorithm~\ref{alg:cap-1}. Its input, $\mathbf{q}_i^{\textit{in}}$, is the output from Algorithm~\ref{alg:cap-22}, as previously applied. Algorithm~\ref{alg:cap-22} is then called again.

The ROI, $\mathbf{v} \leftarrow \mathbf{v}^{\textit{ROI}}$, serves as the visual input (Line~\ref{algo: input}). Attention is computed with respect to entities that have already been identified. Algorithm~\ref{alg:cap-33} governs the decoding and encoding process, which may be called multiple times. Each time, a new label $k$ is generated. These labels could include entities such as \textit{Sparky}, \textit{Dog}, \textit{Black}, and \textit{Happy}.

Due to top-down inference, labels become interdependent. For example, the probability of labeling \textit{Dog} will be influenced by the prior labeling of \textit{Sparky}. In a given scene, multiple labels are generated for each ROI, and the sampling process is repeated over several rounds.

\subsubsection{A Second ROI in a Visual Scene} 


Next, the agent may analyze a second ROI within the context of the scene and the first bounding box. The same sequence of steps—Algorithm~\ref{alg:cap-1}, Algorithm~\ref{alg:cap-22}, and Algorithm~\ref{alg:cap-33}—is applied again. The second ROI might correspond to the entity \textit{Jack}, with additional labels such as \textit{Person}, \textit{Tall}, and \textit{ContentLooking}.

\subsubsection{A Third ROI in a Visual Scene} 

 
 Finally, a third ROI may be defined, encompassing both of the previous bounding boxes. This third ROI could label the relationship between the two entities with a predicate label like \textit{looksAt}, once again utilizing Algorithm~\ref{alg:cap-1}, Algorithm~\ref{alg:cap-22}, and Algorithm~\ref{alg:cap-33}.

\subsubsection{Forming Episodic Indices}


For each perceptual event, an episodic index $t$ is introduced (see Section~\ref{sec:timein}), which can be indexed as an episodic event. Its embedding vector, $\mathbf{a}_t$, is optimized not only to reconstruct the labels of the scene but also to reconstruct the labels for the ROIs, even when the visual inputs are missing. Therefore, the goal is to achieve both symbolic and subsymbolic reconstruction. For further details, see Section~\ref{sec:selfsl}.

\subsubsection{From Labels to Triples}
\label{sec:ftt}

%
 An ROI and an episodic time index serve as unique identifiers (keys) for the other labels associated with an ROI. Assuming that entities are mutually exclusive, one can replace the ROI with the entity it represents, e.g., \textit{Sparky}. The generated labels at time $t$ can then be described as follows
\textit{(Sparky, type, ClassDog)}, 
\textit{(Sparky, type, ColorBlack)}, 
 \textit{(Sparky, type, MoodHappy)}, or more succinctly
 \textit{(Sparky, type, Dog)}, 
 \textit{(Sparky, type, Black)}, and 
\textit{(Sparky, type, Happy)}.\footnote{The relations would be 
 \textit{dog(Sparky)}, 
 \textit{black(Sparky)}, 
 \textit{happy(Sparky)}.}
These are triple statements, and the first entry is the subject, the second is the predicate, and the third is the object.
 Thus, a logical triple statement describes the relationship between concept labels.
 In the context of perception, triples describe currently generated labels, and in the context of episodic memory, past generated labels.
 If an entity's index does not yet exist it can be introduced.

%

One can conceptualize an entity as the object on which a measurement is executed, the predicate describing the type of measurement, and the object representing the outcome of the measurement. Importantly, for reasoning purposes, triples can also be generated with subjects that are not entities, such as
\textit{(Dog, type, Happy)}, 
 \textit{(Black, type, Sparky)}.
 These are referred to as generalized statements in the TB and are useful for embedded symbolic reasoning (see Section~\ref{sec:esr}).

 
 By involving two ROIs (for the subject and object) and an enclosing ROI (for the predicate), triples can also take the form
  \textit{(Sparky, looksAt, Jack)}.

\subsection{Episodic Memory}


According to Tulving, who introduced the term episodic memory~\cite{tulving1985elements}, it stores information about both general and personal events, essentially capturing the information we ``remember.'' It pertains to past observations. In the context of vision, this would involve the reconstruction and decoding of a past scene.

\subsubsection{Episodic Engram}

In the TB framework, an episodic memory engram consists of the episodic index and its associated embedding. When an episodic index is activated, its embedding vector defines the CBS, which then propagates to earlier visual processing layers through embodiment. This process enables the brain to form a subsymbolic understanding of the past event. In the case of vision, this would result in an approximate reconstruction of the past scene. The TB proposes that episodic indices are symbolic in nature, similar to concept indices.

%
%

 \subsubsection{Encoding and Decoding of Episodic Memory}

In the TB approach, episodic memory is realized through the activation of the episodic index corresponding to a past memory. This activation approximately restores the CBS of a past instance.

We begin with Algorithm~\ref{alg:cap-33}, using a neutral input ($\mathbf{q}=0$) and setting 
$k\leftarrow t$ (without sampling). Then, Algorithm~\ref{alg:cap-33} is reapplied to generate labels for the past scene, such as \textit{EnglishGarden} and \textit{Sunny}. Note that if the CBS of the original scene is perfectly restored during episodic recall, the label probabilities will be identical to those at the time of perception.

Next, we consider an ROI from the past scene. The computational evolution neural network of Algorithm~\ref{alg:cap-1} is applied, followed by the application of Algorithm~\ref{alg:cap-33}, which generates labels via sampling. The first label might correspond to the episodic index associated with an ROI in the scene. Labels such as \textit{Sparky}, \textit{Dog}, \textit{Black}, and \textit{Happy} might then be generated. By reapplying Algorithm~\ref{alg:cap-1} and Algorithm~\ref{alg:cap-33}, we can also generate labels for objects and predicates from the past scene.

Essentially, the process of labeling and relationship prediction in episodic memory is similar to that of perception, except that visual inputs are replaced by stored episodic embedding vectors, and attention is not applied. This allows past scenes to be embodied and partially reconstructed. Thus, one can recover not only past memories but also emotions associated with those memories.

As with perception, episodic memory involves exploring different interpretations of a memory. The initiation of an episodic memory and its decoding into symbols, along with the embodiment process, may activate substantial parts of the brain. Based on the generated labels, triple statements can be created.

Episodic memories are not limited to visual events alone. The CBS at any given moment can be stored as an episodic memory with its own unique episodic index. This capability allows for the restoration of previous thoughts, for instance. We will explore this further in Section~\ref{sec:wm}.

\subsubsection{Recent Episodic Memory}

%
%

Recent episodic memory enhances perception by retrieving past perceptual experiences, providing the agent with a sense of the present moment—its current context. To understand its own state and the state of the world, the agent needs to recall recent events, including scenes it has perceived and entities it has encountered. This state information cannot be derived solely from perceptual input.

For example, the agent must remember that, even though current perception offers no indication, it is still in hiding because a bear had been chasing it and might still be lurking outside. In this way, recent episodic memory influences behavior and supports decision-making.

Notably, patients who are unable to form new episodic memories often experience severe deficits in personal orientation and understanding of context. These deficits are commonly linked to significant bilateral damage to the medial temporal lobe (MTL), including the hippocampus~\cite{gluck2013learning}. Age also plays a role in the decline of episodic memory.

\subsubsection{Remote Episodic Memory}

%
%

Remote episodic memory refers to events that are memorable but occurred further in the past. These memories provide valuable decision support by informing the agent about the outcomes of similar past experiences, guiding its behavior based on prior knowledge of what worked or failed. For example, the agent might remember past encounters with {bears} and the dangerous situations that followed, helping it avoid similar risks in the future.

Recall of remote episodic memories is triggered by the similarity between the episodic representation and the current scene representation.

Remote episodic memories can persist for a long time and are associated with brain retraining, which occurs both during conscious activity and sleep~\cite{mcclelland1995there}.

Consider the vast number of movies, TV shows, comic strips, and other media you can recall or at least recognize as either novel or familiar. This illustrates the numerous episodic indices present in the brain, especially if you follow the indexing approach of the TB model.

\subsubsection{Future Episodic Memories and Imagination} 
\label{sec:imagin}

A future episodic memory is a predicted event that, at some point, is expected to become a regular episodic memory. This concept is closely related to prospective memory in cognition.

For example, imagine the agent knows there will be a football game in town this evening, and that the weather will be bad. By creating a future episodic index with labels such as \textit{FootballGame} and \textit{RainyWeather}, the agent might predict bad traffic conditions. This prediction, associated with a future event, is a form of imagination. Importantly, this imagination is grounded and embodied.

Thus, memory guides future behavior through future episodic memories \cite{schacter2012future}. Duncan et al. (2016) describe this process as the integration of relational events by imagining possible future rewards. The value associated with a memory (e.g., reward or threat) is a key aspect of episodic memory. Empirical evidence now strongly supports the role of episodic memory in decision-making tasks \cite{duncan2016memory}.

Consider that the brain has one or more modules evaluating the CBS, which could categorize states as  \textit{PersonallyRewarding}, \textit{ImprovesSocialStatus}, or \textit{Undesirable}. Each state might have an associated index. By imagining different future scenarios based on the agent's potential decisions (e.g., drive to a friend’s house and face bad traffic, or stay home), the agent can assess which decision leads to the maximum expected reward. This facilitates decision support by selecting the action that might lead to the best imagined future scenario.

Imagining future scenarios allows the agent to reason about what is likely to happen under certain conditions, much like logical reasoning.

Perhaps, as many have suggested \cite{dayan1995helmholtz,rao1999predictive,knill2004bayesian,kording2004bayesian,tenenbaum2006theory,griffiths2008bayesian,friston2010free}, the brain functions as a predictive machine. This predictive system may operate as part of the brain's fast-reactive system, also known as implicit memory. For example, it could explain improvements in skills like tennis playing. However, to make predictions over longer time scales, the brain needs to understand the present and relate it to the past—it requires explicit understanding and memory. This is where future episodic memory and imagination play essential roles.

\subsubsection{Temperature Scaling, Simulation, and Bayes Probabilities} 
\label{sec:imagprob}

 To decide on the scenario with the most expected reward, the mind needs to consider the plausibility or likelihood of each scenario. In the TB, only samples are visible, not probabilities. So, how does the brain manage probabilities?

For instance, consider this statement: ``From what I know and from my gut feeling, I bet that \textit{Sparky} will win the race.'' \textit{Sparky} may now be identified as a \textit{RacingDog}. Betting behavior is often cited as evidence that people behave in a Bayes-optimal way, requiring the careful comparison and integration of probabilities.
We can introduce temperature scaling in the $\mathrm{softmax}_{\textit{dom}} $ function ($\mathrm{softmax}_{\textit{dom}} (x)$ becomes $\mathrm{softmax}_{\textit{dom}} (x/T)$ with temperature $T$). By lowering the temperature, we can shift from a ``sample-take-all'' approach to a ``winner-take-all'' approach, where only the most likely outcome is chosen—in this case, \textit{Sparky} wins the race.
Even if the brain cannot always obtain precise probabilities, this method enables optimal decision-making.

Alternatively, as suggested by Sanborn et al. (2016) \cite{sanborn2016bayesian}, the brain may function as a Bayesian sampler. The authors note that ``only with infinite samples does a Bayesian sampler conform to the laws of probability; with finite samples, it systematically generates classic probabilistic reasoning errors, such as the unpacking effect, base-rate neglect, and the conjunction fallacy.''

Statements in semantic memory may also be associated with probabilities (see Section \ref{sec:fcp}), and sampling could be used as a strategy to estimate these probabilities as well.


\subsection{Semantic Memory}
\label{sec:semm}

In cognitive literature, semantic memory refers to the representation of factual knowledge that we know, but cannot recall the specific time or context in which we learned it. It involves statements that are either true, false, or probabilistic in nature. These logical or probabilistic statements are typically time-invariant or change only slowly or infrequently over time.

For example, imagine a friend describes a dog you have never met before. While your episodic memory might provide insights from previous encounters with dogs, your semantic memory likely holds a general understanding of the concept \textit{Dog}—even in the absence of specific episodic recollections.

%

\subsubsection{Semantic Engram}

In the TB framework, a semantic memory engram consists of the concept index $s$ and its corresponding embedding vector $\mathbf{a}_s$. When the embedding of a concept is activated in the representation layer and propagates to earlier layers through embodiment, the brain gains a subsymbolic understanding of that concept.

The embedding vector $\mathbf{a}_s$ can be thought of as a prototype vector, but it exists in the abstract embedding space. In Section~\ref{sec:DNA}, we argue that $\mathbf{a}_s$ acts as the ``signature'' or ``DNA'' of the symbolic index $s$, providing a unique representation of the concept.


\subsubsection{Embedded Semantic Memory}
\label{sec:esm}

%
%

Consider the recall of a symbol $s$, such as \textit{Sparky}, from semantic memory. We begin by applying Algorithm~\ref{alg:cap-33} with a neutral input $\mathbf{q}=0$, where we set $k \leftarrow \bar t$ and do not sample. Here, $\bar t$ represents the index for a time-independent constant embedding vector $\mathbf{\bar a}$.

Next, we apply the computational evolution neural network from Algorithm~\ref{alg:cap-1}, followed by another application of Algorithm~\ref{alg:cap-33}, setting $k \leftarrow s$ and again avoiding sampling. For example, $s$ might be \textit{Sparky}. Repeated applications of Algorithm~\ref{alg:cap-33} might yield labels such as \textit{Dog}, Black, and \textit{Happy}. By reapplying both Algorithm~\ref{alg:cap-1} and Algorithm~\ref{alg:cap-33}, we can also recall relationships, such as \textit{Sparky} is \textit{ownedBy} \textit{Jack}.

Semantic memory recall thus mirrors episodic recall, with the key difference being that $t \leftarrow \bar t$ and $s$ is explicitly set rather than sampled. Importantly, we can use the same (brain) architecture for both semantic and episodic memory retrieval.\footnote{Consider the pCBS. For semantic memory, we need to average over potential visual inputs, i.e., define a prior distribution for the pCBS. We again assume that the prior distribution can be described by independent Bernoulli distributions with pre-activation vector $\mathbf{\bar a}$ with index $\bar t$. This approximates the distribution 
$P^{prior}(X_1, .., X_n) = 
\int P(X_1, ..., X_n | \mathbf{v}) P(\mathbf{v}) d \mathbf{v}$. }

Substituting the expression from Equation~\ref{sec:stateup22} into Equations~\ref{eq:preactact} and \ref{eq:conlabel}, we observe that the sample probability depends on all previous samples. The brain could implement an AND operation across multiple arguments. However, the dependencies are not fully general, which is why the computational evolution neural network, as discussed in Section~\ref{sec:univ}, must be applied to handle more complex dependencies.
%

\subsubsection{Triple-Facts and Conditional Probabilities}
\label{sec:fcp}

%
%
%
%

Perception and episodic memory involve observations. In contrast, semantic memory is concerned with facts. But what exactly constitutes a fact? Consider the perceptual analysis discussed in Section~\ref{sec:ftt}. Suppose each time an ROI is identified as \textit{Sparky} and, if the domain color is determined, it is classified as \textit{Black}. In this case, the statement \textit{(Sparky, type, Black)} is not merely an observation; it is also a fact, independent of any specific episodic instance.

The TB framework can also attach probabilities to facts. For example, assume that whenever \textit{Sparky} appears in a scene, his mood is determined. If \textit{Sparky} is found to be \textit{Happy} 60\% of the time, we can express this fact with the conditional probability:
$P(\textit{Happy} | \textit{Sparky}) = 0.6$. Thus, facts and their associated probabilities can be derived from observational statistics of the generated labels.

\textit{(Sparky, type, Black)} can be interpreted not only as a statement about the color of \textit{Sparky} but also as a rule: ``If an entity is identified as \textit{Sparky}, then it is \textit{Black}.'' This interpretation allows the TB framework to define conditional probabilities for generalized statements like $P(\textit{Black}|\textit{Dog})$. Specifically, the associated probability is determined by the number of times the color is classified as \textit{Black} for an ROI labeled as \textit{Dog}. More explicitly, $P(\textit{Black}|\textit{Sparky})$ can be written as $P(\textit{Color} = \textit{Black}| \textit{Entity} = \textit{Sparky})$ where  \textit{Color} and \textit{Entity} are treated as random variables.

It is important to note that a triple does not necessarily have to be derived from visual perception. For instance, the observation \textit{(Sparky, type, Purebred)} could come from another modality, such as language. In this case, $P(\textit{Purebred} | \textit{Sparky}) = 1$ means that whenever anyone makes a statement about \textit{Sparky}'s pedigree, it is always stated that he is purebred.

Probabilities, however, are complex, especially for the brain. The brain must account for challenges like non-IID (independent and identically distributed) labels, limited sample sizes, and the reliability or uncertainty of the source of the information.

Another issue arises with the mutual exclusivity implied by the softmax normalization over a domain. Consider that \textit{Sparky} is both \textit{Black}  and \textit{White} (checkered). One possible solution would be to introduce a new color category, \textit{BlackWhite}, with $P(\textit{BlackWhite} | \textit{Sparky}) = 1$. Alternatively, perception could label \textit{Sparky} as \textit{Black} 50\% of the time and \textit{White} 50\% of the time, so we would have $P(\textit{Black} | \textit{Sparky}) = P(\textit{White} | \textit{Sparky}) = 0.5$. A third option would be to define separate domains for \textit{Black} and \textit{White}, with labels \textit{Black} and \textit{NotBlack} for the first domain, and \textit{White} and \textit{NotWhite} for the second. In this case, we would get $P(\textit{Black} | \textit{Sparky}) = P(\textit{White} | \textit{Sparky}) = 1$.

This normalization issue may not be as critical for the brain. First, the brain is primarily concerned with generating samples, rather than directly calculating probabilities. Second, the brain may leverage the subsymbolic aspects of semantic or episodic memory recall to conclude that \textit{Sparky} is both \textit{Black} and \textit{White}, without needing to resolve the exact probabilistic distribution. As discussed in Sections~\ref{sec:imagin} and~\ref{sec:imagprob}, the brain may handle probabilities in decision-making scenarios by sampling from potential outcomes.

\subsubsection{Symbolic Semantic Memory}
\label{sec:symreassm}

 
 It is also possible to model semantic memory directly through symbolic methods. Consider the equation for symbolic decoding of semantic memory
 \begin{equation} \label{eq:conlabelsym}
P(Y =k | s) =
 \mathrm{softmax}_{dom} (b_{0, k} + 
 b_{s, k} )
 .
\end{equation}
where  $b_{s, k}$  represents the link between concept $s$ and concept $k$, and $b_{0, k}$ is an offset. In this case, the embeddings of the concepts are not involved. This approach is known as symbolic decoding of semantic memory.

Symbolic decoding is particularly useful for modeling relationships such as successor relations between time indices or \textit{partOf} relationships. For example, it can model the connection between a time index for a scene and the associated ROI, enabling the framework to reason about symbolic relationships between concepts and events.

%

\subsection{Knowledge Graphs}


Triples form a knowledge graph (KG), where concepts are represented as nodes and predicates serve as the links between labels. Some of these triples are certain, in the probabilistic sense described earlier, while others are associated with conditional probabilities. The concept of episodic memory can be linked to a temporal knowledge graph (tKG), capturing the dynamic, time-dependent nature of events and experiences. Generative models for embedded knowledge graphs have also been explored by \cite{loconte2024turn}.

\subsection{Serial Processing as a Prerequisite for Language}

As previously discussed, symbolic and subsymbolic representations are tightly interconnected. For example, one cannot think of \textit{Sparky} without simultaneously activating its subsymbolic embedding and engaging in an embodiment process that might trigger earlier visual and other processing layers in a top-down manner. Similarly, anything represented in the representation layer is constantly decoded into labels (e.g., \textit{Sparky}) through a bottom-up process.

Different brain modules often process information in parallel, particularly during the processing of perceptual pathways. However, symbolic decoding itself follows a serial process, as discussed in Dehaene’s work \cite{dehaene2014consciousness}, where it is described as a ``serial bottleneck'' in consciousness. The importance of sampling in conscious perception is also emphasized, with Dehaene stating that “consciousness is a slow sampler.”

The sequential nature of symbol generation in the TB may have been a key prerequisite for the development of natural language. The natural “outer” language likely stems from an inner, triple-oriented thought language. Individuals largely share common concepts (within the ``inner'' language), even if they do not share a common spoken language (the “outer language”). While embeddings may be unique to each individual brain, recent research suggests they may also be shared to some degree \cite{goldstein2022shared,sucholutsky2023getting}.

Natural language, particularly in its factual aspects, can be seen as a reflection of this inner thought language. For instance, some managers might simply ask, “Just give me the facts.” It has been speculated that even more sophisticated language consists of facts that are connected by fillers—a characteristic observed in certain texts generated by large language models.

Natural (outer) language can also serve as an input to the system. It feeds into the representation layer, where a sentence or paragraph becomes analogous to an episode that is decoded into the inner thought language. Just like any episodic memory with an embedding vector, an input sentence becomes part of the memory system, allowing for recall of both the sentence and its meaning.

Upon completing this manuscript, we noticed interesting parallels with the discussion in \cite{wong2023word}. Our inner thought language can be linked to their concept of a probabilistic language of thought (PLoT), which is rooted in probabilistic programming and includes domain knowledge that can be learned from language by large language models. We agree with the paper's argument that PLoT (and our inner thought language) likely preceded natural language, both in child development and in evolution, with natural language evolving on this foundation. While the paper focuses on System 2 functionalities, most operations in the TB align with the effortless, intuitive processes of System 1.

\section{Self-supervised Learning}
\label{sec:selfsl}

\subsection{Associative Learning?}

Consider simple associations between indices and embeddings: an episodic memory embedding can be formed by storing the CBS at the time of perception. Similarly, the embedding vector for a concept could be the average of all CBSs where the concept appeared. This approach might have served as the foundation of biological evolution. The TB, however, employs self-supervised learning, which is described in the following sections and leads to improved performance.


\subsection{Self-generated Labels}

In this model, no external agent provides training data or symbolic labels. Instead, perception itself generates labels through stochastic sampling, which the brain then treats as actual labels for training. This is a form of bootstrap learning, where the model's predictions are used as targets for further learning. See the bootstrap Widrow-Hoff rule \cite{hinton1990bootstrap} and learning with pseudo labels \cite{lee2013pseudo} for related concepts. Self-generated labels are incorporated into various cost functions, as discussed in \cite{bengio2015towards}, which outlines different principled methods for self-supervised adaptation in deep learning.


\subsection{Embedding Learning with Self-generated Labels }
\label{sec:self2}

%
%

Let’s assume that a scene, $\mathbf{v}^{\textit{scene}}$, has been observed, and the focus shifts to a region of interest (ROI) with $\mathbf{v}^{\textit{ROI}}$. For simplicity, we exclude attention mechanisms. Assume that the labels $s = \textit{Sparky}$ and $k = \textit{Dog}$ are generated.

For perception, the contribution to the log-likelihood (negative cost function) is
$
\log P(s | \mathbf{v}^{\textit{ROI}}, \mathbf{v}^{\textit{scene}} ) 
+ \log P(k | s, \mathbf{v}^{\textit{ROI}}, \mathbf{v}^{\textit{scene}} )
$ which becomes
\[
 \log \mathrm{softmax}_{\textit{dom}} ( \mathbf{a}^{\top}_s 
\mathrm{sig}( \mathbf{h}_1))
+ 
\log \mathrm{softmax}_{\textit{dom}} 
( \mathbf{a}^{\top}_k 
\mathrm{sig}(\mathbf{a}_s + \mathbf{h}_1)) .
\]
where $\mathbf{h}_1 = \mathbf{g}(\mathbf{v}^{\textit{ROI}}) + \mathbf{g}(\mathbf{v}^{\textit{scene}}) + \mathbf{f}^{\textit{NN}}(\mathbf{g}(\mathbf{v}^{\textit{scene}}))$.

Let’s assume this is the first time the label $s = \textit{Sparky}$ has been observed. The first term in the cost function will attempt to align the embedding $\mathbf{a}_s$ with the visual input. The second cost term will attempt to align $\mathbf{a}_s$ with the embedding for \textit{Dog}, thereby integrating the “dogginess” into \textit{Sparky}'s embedding.

For the episodic memory at time $t$, the contribution to the log-likelihood is
%
%
%
 $\log P(s| t) + \log 
P(k | s, t)$ which becomes
\[
 \log \mathrm{softmax}_{\textit{dom}} ( \mathbf{a}^{\top}_s 
\mathrm{sig}(\mathbf{h}_2))
+ \log \mathrm{softmax}_{\textit{dom}} 
( \mathbf{a}^{\top}_k 
\mathrm{sig}(\mathbf{a}_s + \mathbf{h}_2)) .
\]
where $\mathbf{h}_2 = \mathbf{a}_t + \mathbf{f}^{\textit{NN}}(\mathbf{a}_t)$.\footnote{If the embedding vector for the ROI, $\mathbf{a}_{t_{roi}}$, is available, $\mathbf{h}_2$ can be updated to $\mathbf{h}_2 = \mathbf{a}_{t_{roi}} + \mathbf{a}_t + \mathbf{f}^{\textit{NN}}(\mathbf{a}_t)$.}

For semantic memory related to entity $s$, the contribution to the log-likelihood is:
%
%
\[
\log P(k | s)
= 
 \log \mathrm{softmax}_{\textit{dom}} 
( \mathbf{a}^{\top}_k 
 \mathrm{sig}(\mathbf{a}_s + \mathbf{h}_3 )) 
\]
where $\mathbf{h}_3 = \mathbf{\bar a} + \mathbf{f}^{\textit{NN}}(\mathbf{\bar a})$. This cost term aims to align the embedding of \textit{Sparky} with the embedding of \textit{Dog}. As a result, when semantic memory recalls \textit{Sparky}, it is likely that \textit{Dog} will be sampled next. The embedding of the index \textit{Sparky} will integrate all labels generated in context with \textit{Sparky}.

We can also derive cost terms for triples with predicates other than \textit{type}, such as \textit{(Sparky, looksAt, Jack)}.\footnote{Note that the equations are not commutative: Sampling $s$ first and then $k$ results in a different cost compared to sampling $k$ first and then $s$.}

A consolidated episodic memory recall is more consistent with the agent's worldview. Past memories are slightly adjusted to better fit the agent’s expectations. Memories that align with the agent's understanding are more easily recalled than those from unfamiliar or foreign contexts. As discussed in \cite{craik1975depth,kafkas2018expectation}, humans can only effectively report past scenes when they align with the agent’s worldview.

The adaptation of embedding vectors and the formation of semantic memory likely occur during the memory consolidation process, which is believed to take place during sleep.

%
%
%
%
%
%
%

\subsection{Autoencoder Learning}
\label{sec:self}

In Section \ref{sec:auto}, we discussed how scene input can be used to train an autoencoder, a key component in realizing embodiment. The cost functions derived from a perceptual autoencoder can help regularize and stabilize the learning process.


\subsection{Memory Recall Changes Memories}
\label{sec:falsem}

As previously mentioned, the recall of episodic or semantic memories involves the complex decoding of embedding vectors, during which samples are generated. The brain may then decide to train on these samples, providing an explanation for the well-documented phenomenon that memory traces can change upon retrieval. In fact, memories can even be manipulated, as demonstrated in the work of Loftus \cite{loftus1991witness,loftus2019human}.


\subsection{New Indices}

For existing indices, predicted labels can be used to adapt their embeddings. However, the brain also needs to create new indices. A new perceptual experience is represented by a newly formed episodic index and its corresponding learned embedding. At a slower rate, new entity indices are introduced, and even more slowly, new indices for classes and attributes are required. The neurobiological aspects of forming new indices are extensively discussed in \cite{tresp2023tensor}.


%
%
 
\section{Multitasking, Cognitive Control, and Working Memory}
\label{sec:wm}

\subsection{Multitasking by Multiplexing}

Consider the scenario where an agent is planning a friend's visit while simultaneously driving a car. Suddenly, the traffic situation demands full attention for a few seconds, after which the agent resumes planning the visit. How can the brain manage this?

One leading hypothesis suggests that it only appears as if the mind can think of several things at once. In reality, at the conscious level, we rapidly switch between mental states. This implies that cognitive attention may represent a serial bottleneck in the brain. An extreme example of this phenomenon is multitasking, where humans seem to manage two tasks at once, such as driving and conversing. However, cognitive multitasking is an illusion—the brain performs multiplexing, rather than truly multitasking \cite{miller2015working}.

The TB model supports the concept of multitasking via multiplexing. In the example provided, the evening-planning CBS is stored as the embedding vector of an episodic index. After the traffic situation is resolved, this CBS is retrieved by activating the corresponding episodic index.

However, multitasking by multiplexing might have a capacity limit. The capacity of working memory is often estimated to be seven items or fewer \cite{miller1956magical,engle2001working,cowan2010magical}, meaning the mind can switch between seven or fewer mental states. In the context of the TB, cognitive control likely involves managing up to seven episodic or concept indices and their embeddings for problem-solving.

A specific example of multitasking is the use of episodic and semantic memory during perception. If time allows, the brain can activate these memory systems to supplement ongoing perceptual activity with background information, such as recent episodic memory (providing state information) or remote episodic memory (guiding decision-making). See Figure~\ref{fig:TB-dogench}. The multitasking between perception and memory is easily supported by the TB model but may be more challenging for other technical models.

Samples generated by semantic memory can be considered as prior samples, while samples generated from perception or episodic memory can be seen as data. In \cite{tresp2023tensor}, this relationship was explored in the context of Dirichlet fusion.

\subsection{Cognitive Control and Working Memory}

Many mental tasks appear to require some form of cognitive control. Multitasking is an obvious example: How does the brain decide which mental or physical task to prioritize at any given moment?

Another example is the navigation of semantic memory. Imagine the agent is thinking about \textit{Sparky}. Should the mind remain focused on \textit{Sparky}, or should it digress? For instance, thinking of \textit{Sparky} might lead to recalling \textit{Jack} (his owner), who recently married \textit{Mary}, who lives in \textit{Munich}. This might trigger thoughts of the \textit{English Garden} in Munich, known for its charming \textit{BeerGardens}, which in turn reminds the agent to buy some beer on the way home.\footnote{This type of digression is humorously depicted in the character J.D. from the sitcom Scrubs.}

A third example involves the sequential selection of regions of interest (ROIs) in visual processing, as explored in \cite{sabatinelli2014timing}.

Cognitive control appears to be a central faculty of consciousness and conscious decision-making. But who—or what—is ultimately in charge? Technically, the module governing cognitive control may be optimized through reinforcement learning, guiding decisions and actions based on learned priorities and rewards.

%
%
%

 \begin{figure}[t]
\begin{center}
 \includegraphics[width=0.8\linewidth]{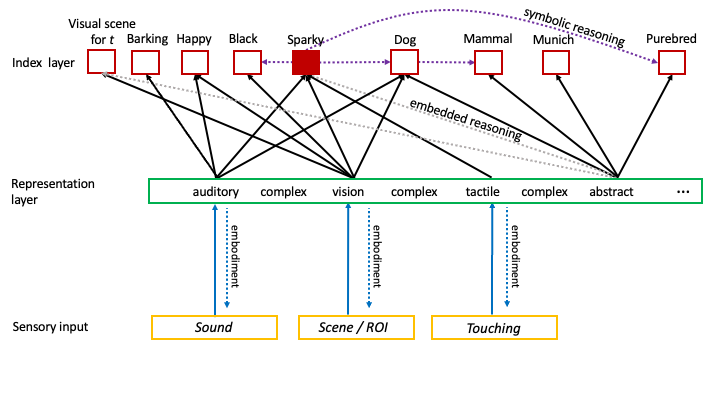}
 \includegraphics[width=0.8\linewidth]{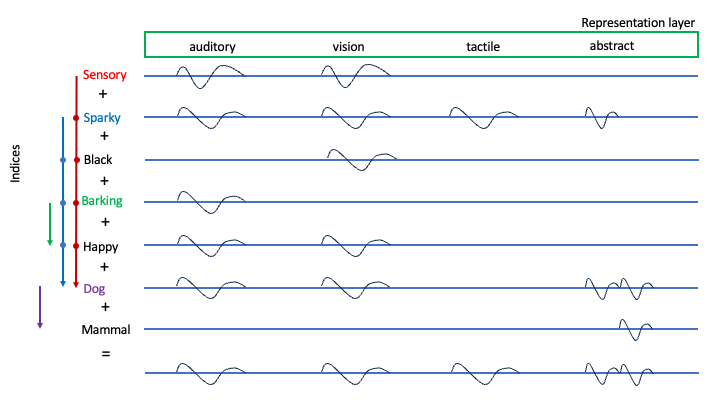}
\end{center}
 \caption{
{\color{black} Multimodality and Reasoning. Top: Some dimensions in the representation layer are visual, some auditory, some tactile, and some abstract. 
Different dimensions communicate with different perceptual paths and 
 symbolic indices. The horizontal dotted lines between the indices indicate symbolic semantic memory and reasoning. 
 Bottom: Sequential operations in perception. 
 Indices are sampled and added from top to bottom. 
 The first line indicates that sensory input activates auditory and visual dimensions of the representation layer. 
 Then the index \textit{Sparky} fires which adds information on auditory and visual dimensions and in addition on tactile and abstract dimensions. 
 The decoding continues with \textit{Black}, \textit{Barking}, \textit{Happy}, \textit{Dog} and \textit{Mammal}. 
 \textit{Mammal} fires because \textit{Dog} has fired; there is no direct overlap in the representation layer with \textit{Sparky}. This can be achieved by 
 an adaptation of \textit{Sparky}'s embedding (embedded reasoning).
 In an episodic memory, the first line is represented by an embedding vector. 
 In semantic memory, the first line is missing, and the process starts by activating the index for \textit{Sparky}. 
 }
 }
\label{fig:TB-segm}
\end{figure}

\section{Multimodality, Generalization, Reasoning, and Semantics}
\label{sec:reason}

Is the TB merely a large database, as might be inferred from its focus on past observations and statistical patterns? Certainly not. While concept labels sampled from past experiences are not stored explicitly, they serve to refine perception and enable generalization in memory through the learning and fine-tuning of embedding vectors, as we discuss in this section. The TB clearly generalizes to new perceptual inputs, such as novel scenes. Since labels are self-generated, there is no traditional training data to overfit to—another indication of its capacity for generalization. Let us now explore reasoning, which can be best understood through the lens of multimodality.


\subsection{Multimodality: More Dimensions} 
\label{sec:mulimodality}

The TB is inherently multimodal by design. Multiple modules communicate with the representation layer, especially sensory modalities. Each modality contributes to specific dimensions of the representation layer. For example, some dimensions may predominantly process visual inputs, while others are auditory, tactile, or abstract, as illustrated in Figure \ref{fig:TB-segm}.

Embeddings may also be sparse. For instance, the index \textit{Black} primarily interacts with visual dimensions, while \textit{Loud} or \textit{Barking} interacts with auditory dimensions. However, concepts sometimes transcend modalities—for example, one might describe an image as ``loud.''

Consider the entity \textit{Sparky}. Visual dimensions provide information about his appearance, such as color and shape. Auditory dimensions describe his bark and other sounds, tactile dimensions represent how he feels, and abstract dimensions capture higher-level associations. This rich, multimodal inference process enables the TB to construct a holistic understanding of \textit{Sparky}, as depicted in Figure \ref{fig:TB-segm}.

%
%

\subsection{Embedded Symbolic Reasoning and Chaining}
\label{sec:esr}

Consider the example in Figure \ref{fig:TB-segm}, where sensory input activates both visual and auditory dimensions of the representation layer. This bottom-up inference might sample the index \textit{Sparky}, activating all regions of the representation layer associated with him, including visual, auditory, tactile, and abstract dimensions.

The activation of \textit{Sparky} supports the activation of \textit{Dog}. Once \textit{Dog} is activated, \textit{Mammal} is triggered, demonstrating symbolic chaining. Notably, \textit{Mammal} does not directly connect to the visual dimensions of \textit{Sparky}, but rather through its association with \textit{Dog}. This process illustrates the brain’s capacity for symbolic embedded reasoning—concluding that \textit{Sparky} is a \textit{Mammal} without directly modifying his embedding.


\subsection{Embedded Reasoning and Materialization}
\label{sec:embr}

Now consider the same scenario, but assume the embedding for \textit{Sparky} has been updated to integrate the label \textit{Mammal}. In this case, activating \textit{Sparky}'s index might directly sample \textit{Mammal}, bypassing the intermediate step of activating \textit{Dog}.

This form of reasoning, akin to that used in recommendation systems~\cite{koren2009matrix}, resembles the process of materialization in logic and database theory, where precomputed relationships facilitate faster, direct retrieval.


\subsection{Embedded Reasoning Creates Episodic Similarities}

Imagine the agent has stored two scenes: one from the \textit{English Garden} at time 
$t_1$, labeled with \textit{Sunny}, and another from \textit{Marienplatz} at time 
$t_2$, labeled with \textit{Rainy}. Initially, their episodic embeddings and visual labels may share little similarity.

However, a location module in the brain might associate both scenes with \textit{Munich}. This location-generated label can be shared across both episodes. As the episodic embeddings are updated, they begin to integrate the embedding for \textit{Munich}. For instance, the embedding vector for 
$t_1$   is adapted to generate the label \textit{Munich}, as is the embedding for 
$t_2$.

As a result, the two engrams become similar in their embeddings. Recalling one episodic memory may activate the other, and recalling the semantic concept \textit{Munich} could now activate both episodic memories.

This is a powerful example of how perceptual and symbolic embedding adaptation can enrich memory and reasoning.

%
%
%
%


\subsection{Symbolic Reasoning}
\label{sec:symbr}

Symbolic reasoning refers to reasoning processes that do not explicitly utilize the embeddings of symbolic indices. As discussed in Section~\ref{sec:symreassm}, symbolic decoding in semantic memory relies solely on symbolic indices, independent of their embeddings.

This type of reasoning can represent binary relationships, such as the association between \textit{Dog} and \textit{Black}, and supports chaining operations. For example, if \textit{Sparky} is sampled in a scene, triggering the activation of \textit{Dog}, which in turn activates \textit{Mammal}, the mind can infer that \textit{Sparky} is a \textit{Mammal}.

Chaining can be materialized by introducing a direct excitatory connection from \textit{Sparky} to \textit{Mammal}, bypassing intermediate steps. More complex forms of reasoning—such as those based on relational Bayesian networks—are also conceivable within the framework of symbolic indices. Symbolic reasoning has been implemented as probabilistic programming, as demonstrated in \cite{wong2023word}.


\subsection{Universal Approximation Property}
\label{sec:univ}

In Section~\ref{sec:esm}, we noted that a sample can depend nonlinearly on prior samples, though this dependency is not fully general. To achieve complete generality, the involvement of the evolution neural network is necessary.

Consider an example where an entity serves as both a subject and an object, such as \textit{Sparky}. Sampling the entity in its role as a subject might yield attributes like \textit{Dog} and \textit{Black}, while sampling it as an object might result in \textit{Cute}. The latter typically exhibits a nonlinear dependency on the former because the evolution neural network governs the transition from subject to object roles.

This process enables the brain to leverage the evolution neural network for general mappings, thereby achieving a universal approximation capability.


\subsection{The Embedding is the ``DNA''}
\label{sec:DNA}

What encapsulates the semantics of a concept like \textit{Sparky}? The embedding of \textit{Sparky} serves as its unique signature or ``DNA.'' Decoding this ``DNA'' involves activating the representation layer with \textit{Sparky}'s embedding, which informs various brain modules about his sensory and other properties.

For instance, decoding conveys how \textit{Sparky} looks, sounds, and feels. Semantic memory further provides information about \textit{Sparky}'s attributes, such as being \textit{Black}, as well as his relationships, like being owned by \textit{Jack} and loved by \textit{Mary}. Past episodic memories involving \textit{Sparky} may also be recalled, contributing additional context.

Through this process, substantial regions of the brain’s sensory and memory systems collaboratively ``decode'' \textit{Sparky}'s embedding. This adaptable signature is unique to the individual brain interpreting it. Consequently, thinking intensely about \textit{Sparky} triggers a ``firework'' of neural activity—a phenomenon mirrored during perception and episodic memory recall.

%

\subsection{Working Memory and Cognitive Control}

Working memory and cognitive control are often seen as hallmarks of intelligence. As discussed in Section~\ref{sec:wm}, working memory may function by manipulating a limited number of symbolic indices to guide decision-making.

While perception and memory provide foundational inputs, reasoning also relies on model-based inference, as elaborated in \cite{wong2023word}. Human intelligence, however, is multifaceted and depends on a diverse array of functional modules working in concert.


\section{Summary and Conclusions}
\label{sec:summ}

Perception and memory recall are processes that engage substantial regions of the brain. The Tensor Brain (TB) framework proposes that symbolic indices are activated through a bottom-up process in both functions. In turn, these activated indices communicate with the representation layer and earlier processing layers in a top-down manner. During perception and memory recall, multiple indices are sequentially activated, creating a dynamic interaction. This interplay can be likened to a``dance,'' where symbolic indices activate subsymbolic representation layers and are themselves influenced in return.

A key innovation of the TB is learning through self-generated labels, a concept that has shown promising initial results. However, it carries inherent risks. As discussed, not only does the formation of new memories generate labels, but so does memory retrieval. The brain may use these labels for adaptation, which could lead to potential falsification or alteration of past memories.

The brain has often been described as a prediction machine. Over shorter time scales, this predictive mechanism may integrate with the brain’s fast, reactive systems and implicit memory. However, making predictions on the longer time scales relevant to the TB requires more than reactivity; it necessitates explicit understanding and explicit memory. To predict effectively, the mind must connect the present to both the past and the future. This includes imagining future possibilities—a process requiring memory-based imagination, supplemented by model-based imagination.

The TB framework demonstrated embedded and symbolic reasoning, and explored their relationship with semantic memory. It also examined the roles of working memory and cognitive control. While perception and memory may provide the foundation for these functions, more complex forms of reasoning appear to be uniquely human. It has been speculated that language plays a crucial role, as both reasoning and language rely on distinct but interacting brain regions.

Despite these advancements, it is important to remember that the brain operates as a multitasking system. Most of its activities occur outside conscious awareness, employing diverse strategies to address specific problems concurrently.

The debate over localized versus distributed representations in the brain remains active. Evidence supports the existence of concept-specific representations. In particular, concept cells have been identified in the medial temporal lobe (MTL). These neurons exhibit remarkable selectivity, responding to a wide range of representations for specific individuals, landmarks, or objects—and even to textual representations such as letter strings corresponding to their names \cite{quiroga2012concept,quiroga2005invariant}. This lends support to  index-specific ``symbolic'' representations.

A distinctive feature of the TB is its reliance on the sampling of indices. But what initiates this sampling process? The nature of this initiation bears intriguing parallels to the collapse of the wave function in quantum mechanics, triggered by measurement. These questions are explored further in \cite{tresp2024QTB}.

 \bibliography{HBQuantumTensorBrainSem}{}

\begin{thebibliography}{10}

\bibitem{baars1997theater}
Bernard~J Baars.
\newblock {\em In the theater of consciousness: The workspace of the mind}.
\newblock Oxford University Press, USA, 1997.

\bibitem{baier2017improving}
Stephan Baier, Yunpu Ma, and Volker Tresp.
\newblock Improving visual relationship detection using semantic modeling.
\newblock In {\em {ISWC}}, volume~16. Springer, 2017.

\bibitem{bengio2015towards}
Yoshua Bengio, Dong-Hyun Lee, Jorg Bornschein, Thomas Mesnard, and Zhouhan Lin.
\newblock Towards biologically plausible deep learning.
\newblock {\em arXiv preprint arXiv:1502.04156}, 2015.

\bibitem{bordestranslating2013}
Antoine Bordes, Nicolas Usunier, Alberto Garcia-Duran, Jason Weston, and Oksana
  Yakhnenko.
\newblock Translating {Embeddings} for {Modeling} {Multi}-relational {Data}.
\newblock In {\em Advances in {Neural} {Information} {Processing} {Systems}
  26}, 2013.

\bibitem{cowan2010magical}
Nelson Cowan.
\newblock The magical mystery four: How is working memory capacity limited, and
  why?
\newblock {\em Current directions in psychological science}, 19(1):51--57,
  2010.

\bibitem{craik1975depth}
Fergus~IM Craik and Endel Tulving.
\newblock Depth of processing and the retention of words in episodic memory.
\newblock {\em Journal of experimental Psychology: general}, 104(3):268, 1975.

\bibitem{dayan1995helmholtz}
Peter Dayan, Geoffrey~E Hinton, Radford~M Neal, and Richard~S Zemel.
\newblock The {H}elmholtz machine.
\newblock {\em Neural computation}, 7(5):889--904, 1995.

\bibitem{dehaene2014consciousness}
Stanislas Dehaene.
\newblock {\em Consciousness and the brain: Deciphering how the brain codes our
  thoughts}.
\newblock Penguin, 2014.

\bibitem{duncan2016memory}
Katherine~D Duncan and Daphna Shohamy.
\newblock Memory states influence value-based decisions.
\newblock {\em Journal of Experimental Psychology: General}, 145(11):1420,
  2016.

\bibitem{engle2001working}
Randall~W Engle.
\newblock What is working memory capacity?
\newblock In H.~L. Roediger, III, I.~Neath J.~S.~Nairne, and A.~M. Surprenant,
  editors, {\em The nature of remembering: Essays in honor of {R}obert {G.}
  {C}rowder}. American Psychological Association Press, 2001.

\bibitem{friston2010free}
Karl Friston.
\newblock The free-energy principle: a unified brain theory?
\newblock {\em Nature Reviews Neuroscience}, 11(2):127--138, 2010.

\bibitem{gluck2013learning}
Mark~A Gluck, Eduardo Mercado, and Catherine~E Myers.
\newblock {\em Learning and memory: From brain to behavior}.
\newblock Palgrave, 2013.

\bibitem{goldstein2022shared}
Ariel Goldstein, Zaid Zada, Eliav Buchnik, Mariano Schain, Amy Price, Bobbi
  Aubrey, Samuel~A Nastase, Amir Feder, Dotan Emanuel, Alon Cohen, et~al.
\newblock Shared computational principles for language processing in humans and
  deep language models.
\newblock {\em Nature neuroscience}, 25(3):369--380, 2022.

\bibitem{griffiths2008bayesian}
Thomas~L Griffiths, Charles Kemp, and Joshua~B Tenenbaum.
\newblock Bayesian models of cognition.
\newblock In {\em The Cambridge Handbook of Computational Psychology}.
  Cambridge University Press, 2008.

\bibitem{he2016deep}
Kaiming He, Xiangyu Zhang, Shaoqing Ren, and Jian Sun.
\newblock Deep residual learning for image recognition.
\newblock In {\em Proceedings of the IEEE conference on computer vision and
  pattern recognition}, 2016.

\bibitem{hinton1990bootstrap}
Geoffrey~E Hinton and Steven~J Nowlan.
\newblock The bootstrap {W}idrow-{H}off rule as a cluster-formation algorithm.
\newblock {\em Neural Computation}, 2(3):355--362, 1990.

\bibitem{hochreiter1997long}
Sepp Hochreiter and J{\"u}rgen Schmidhuber.
\newblock Long short-term memory.
\newblock {\em Neural computation}, 9(8):1735--1780, 1997.

\bibitem{johnson2015image}
Justin Johnson, Ranjay Krishna, Michael Stark, Li-Jia Li, David Shamma, Michael
  Bernstein, and Li~Fei-Fei.
\newblock Image retrieval using scene graphs.
\newblock In {\em CVPR}, 2015.

\bibitem{kafkas2018expectation}
Alex Kafkas and Daniela Montaldi.
\newblock Expectation affects learning and modulates memory experience at
  retrieval.
\newblock {\em Cognition}, 180:123--134, 2018.

\bibitem{knill2004bayesian}
David~C Knill and Alexandre Pouget.
\newblock The {B}ayesian brain: the role of uncertainty in neural coding and
  computation.
\newblock {\em Trends in Neurosciences}, 27(12):712--719, 2004.

\bibitem{kording2004bayesian}
Konrad~P K{\"o}rding, Shih-pi Ku, and Daniel~M Wolpert.
\newblock Bayesian integration in force estimation.
\newblock {\em Journal of Neurophysiology}, 92(5):3161--3165, 2004.

\bibitem{koren2009matrix}
Yehuda Koren, Robert Bell, and Chris Volinsky.
\newblock Matrix factorization techniques for recommender systems.
\newblock {\em Computer}, 42(8):30--37, 2009.

\bibitem{lee2013pseudo}
Dong-Hyun Lee et~al.
\newblock Pseudo-label: The simple and efficient semi-supervised learning
  method for deep neural networks.
\newblock In {\em Workshop on challenges in representation learning, ICML},
  2013.

\bibitem{loconte2024turn}
Lorenzo Loconte, Nicola Di~Mauro, Robert Peharz, and Antonio Vergari.
\newblock How to turn your knowledge graph embeddings into generative models.
\newblock {\em Advances in Neural Information Processing Systems}, 36, 2024.

\bibitem{loftus1991witness}
Elizabeth~F Loftus and Katherine Ketcham.
\newblock {\em Witness for the defense: The accused, the eyewitness, and the
  expert who puts memory on trial}.
\newblock Macmillan, 1991.

\bibitem{loftus2019human}
Geoffrey~R Loftus and Elizabeth~F Loftus.
\newblock {\em Human memory: The processing of information}.
\newblock Psychology Press, 2019.

\bibitem{ma2018embedding}
Yunpu Ma, Volker Tresp, and Erik~A Daxberger.
\newblock Embedding models for episodic knowledge graphs.
\newblock {\em Journal of Web Semantics}, 2018.

\bibitem{mcclelland1995there}
James~L McClelland, Bruce~L McNaughton, and Randall~C O'Reilly.
\newblock Why there are complementary learning systems in the hippocampus and
  neocortex: insights from the successes and failures of connectionist models
  of learning and memory.
\newblock {\em Psychological review}, 102(3):419, 1995.

\bibitem{mikolov2013distributed}
Tomas Mikolov, Ilya Sutskever, Kai Chen, Greg~S Corrado, and Jeff Dean.
\newblock Distributed representations of words and phrases and their
  compositionality.
\newblock {\em Advances in neural information processing systems}, 26, 2013.

\bibitem{miller2015working}
Earl~K Miller and Timothy~J Buschman.
\newblock Working memory capacity: Limits on the bandwidth of cognition.
\newblock {\em Daedalus}, 144(1):112--122, 2015.

\bibitem{miller1956magical}
George~A Miller.
\newblock The magical number seven, plus or minus two: Some limits on our
  capacity for processing information.
\newblock {\em Psychological review}, 63(2):81, 1956.

\bibitem{quiroga2005invariant}
R~Quian Quiroga, Leila Reddy, Gabriel Kreiman, Christof Koch, and Itzhak Fried.
\newblock Invariant visual representation by single neurons in the human brain.
\newblock {\em Nature}, 435(7045):1102--1107, 2005.

\bibitem{quiroga2012concept}
Rodrigo~Quian Quiroga.
\newblock Concept cells: the building blocks of declarative memory functions.
\newblock {\em Nat Rev Neurosci}, 13(8), 2012.

\bibitem{rao1999predictive}
Rajesh~PN Rao and Dana~H Ballard.
\newblock Predictive coding in the visual cortex: a functional interpretation
  of some extra-classical receptive-field effects.
\newblock {\em Nature neuroscience}, 2(1):79--87, 1999.

\bibitem{sabatinelli2014timing}
D~Sabatinelli, DW~Frank, TJ~Wanger, M~Dhamala, BM~Adhikari, and X~Li.
\newblock The timing and directional connectivity of human frontoparietal and
  ventral visual attention networks in emotional scene perception.
\newblock {\em Neuroscience}, 277:229--238, 2014.

\bibitem{sanborn2016bayesian}
Adam~N Sanborn and Nick Chater.
\newblock {B}ayesian brains without probabilities.
\newblock {\em Trends in cognitive sciences}, 20(12):883--893, 2016.

\bibitem{schacter2012future}
Daniel~L Schacter, Donna~Rose Addis, Demis Hassabis, Victoria~C Martin,
  R~Nathan Spreng, and Karl~K Szpunar.
\newblock The future of memory: remembering, imagining, and the brain.
\newblock {\em Neuron}, 76(4):677--694, 2012.

\bibitem{sharifzadeh2021classification}
Sahand Sharifzadeh, Sina~Moayed Baharlou, and Volker Tresp.
\newblock Classification by attention: Scene graph classification with prior
  knowledge.
\newblock In {\em AAAI}, volume~35, 2021.

\bibitem{sucholutsky2023getting}
Ilia Sucholutsky, Lukas Muttenthaler, Adrian Weller, Andi Peng, Andreea Bobu,
  Been Kim, Bradley~C Love, Erin Grant, Jascha Achterberg, Joshua~B Tenenbaum,
  et~al.
\newblock Getting aligned on representational alignment.
\newblock {\em arXiv preprint arXiv:2310.13018}, 2023.

\bibitem{tenenbaum2006theory}
Joshua~B Tenenbaum, Thomas~L Griffiths, and Charles Kemp.
\newblock Theory-based {B}ayesian models of inductive learning and reasoning.
\newblock {\em Trends in cognitive sciences}, 10(7):309--318, 2006.

\bibitem{teyler1986hippocampal}
Timothy~J Teyler and Pascal DiScenna.
\newblock The hippocampal memory indexing theory.
\newblock {\em Behavioral neuroscience}, 1986.

\bibitem{tresp2015learning}
Volker Tresp, Crist{\'o}bal Esteban, Yinchong Yang, Stephan Baier, and Denis
  Krompa{\ss}.
\newblock Learning with memory embeddings.
\newblock {\em arXiv preprint arXiv:1511.07972}, 2015.

\bibitem{tresp2024QTB}
Volker Tresp and Hang Li.
\newblock {B}ayes or {H}eisenberg: Who(se) rules?
\newblock {\em arXiv preprint}, 2024.

\bibitem{tresp2017embedding}
Volker Tresp, Yunpu Ma, Stephan Baier, and Yinchong Yang.
\newblock Embedding learning for declarative memories.
\newblock In {\em European Semantic Web Conference}, volume~14. Springer, 2017.

\bibitem{tresp2020tensor}
Volker Tresp, Sahand Sharifzadeh, Dario Konopatzki, and Yunpu Ma.
\newblock The tensor brain: Semantic decoding for perception and memory.
\newblock {\em arXiv preprint arXiv:2001.11027}, 2020.

\bibitem{tresp2023tensor}
Volker Tresp, Sahand Sharifzadeh, Hang Li, Dario Konopatzki, and Yunpu Ma.
\newblock The tensor brain: A unified theory of perception, memory, and
  semantic decoding.
\newblock {\em Neural Computation}, 35(2):156--227, 2023.

\bibitem{tulving1985elements}
Endel Tulving.
\newblock {\em Elements of episodic memory}.
\newblock Oxford University Press, 1985.

\bibitem{wong2023word}
Lionel Wong, Gabriel Grand, Alexander~K Lew, Noah~D Goodman, Vikash~K
  Mansinghka, Jacob Andreas, and Joshua~B Tenenbaum.
\newblock From word models to world models: Translating from natural language
  to the probabilistic language of thought.
\newblock {\em arXiv preprint arXiv:2306.12672}, 2023.

\end{thebibliography}
\bibliographystyle{plain}

 \end{document}